\documentclass[11pt]{article}
\usepackage{acl2014}
\usepackage{times}
\usepackage{url}
\usepackage{multirow}
\usepackage{tabularx}
\usepackage{tabulary}
\usepackage{latexsym}

\usepackage{graphicx}
\graphicspath{ {images/} }

\usepackage{comment}
\usepackage{CJKutf8}
\usepackage{hyperref}
\usepackage{hhline}

\title{A Hybrid Word-Character Model for Abstractive Summarization}

\author{Chieh-Teng Chang, Chi-Chia Huang, Chih-Yuan Yang \and Jane Yung-Jen Hsu \\
\mbox{}\\
Department of Computer Science and Information Engineering \\
National Taiwan University, Taipei, Taiwan \\
\{scott820914, chifatty\}@gmail.com\\
\{yangchihyuan, yjhsu\}@csie.ntu.edu.tw}

\date{}

\begin{document}
\begin{CJK*}{UTF8}{gbsn}

\maketitle
\begin{abstract}

Automatic abstractive text summarization is an important and challenging research topic of natural language processing. Among many widely used languages, the Chinese language has a special property that a Chinese character contains rich information comparable to a word. Existing Chinese text summarization methods, either adopt totally character-based or word-based representations, fail to fully exploit the information carried by both representations. To accurately capture the essence of articles, we propose a hybrid word-character approach (HWC) which preserves the advantages of both word-based and character-based representations. We evaluate the advantage of the proposed HWC approach by applying it to two existing methods, and discover that it generates state-of-the-art performance with a margin of 24 ROUGE points on a widely used dataset LCSTS. In addition, we find an issue contained in the LCSTS dataset and offer a script to remove overlapping pairs (a summary and a short text) to create a clean dataset for the community. The proposed HWC approach also generates the best performance on the new, clean LCSTS dataset.

\end{abstract}

\section{Introduction}

Text summarization aims to create a short, fluent, and effective summary from a long text document. Since the rapid growth of information stored in the textual form in digital documents, summaries greatly help address the amount of text data available online for searching useful information and consuming relevant articles.
Text summarization covers various types. General summarization improves the effectiveness of indexing and reduces the bias caused by humans. Query-focused summarization takes preferences into consideration to satisfy individual needs of information. Multiple-document summarization aims to generate summaries across multiple documents about the same topic. Extractive summarization combines a group of informative pieces of text from the source without changing them, and abstractive summarization generates entirely new sentences by absorbing the information contained in the source text~\cite{TorresMoreno2014,Gambhir2017}.

The recently rapid development of neural networks brings significant advances into text summarization, especially from the family of attentional sequence to sequence (Seq2Seq) architectures~\cite{sutskever2014seq2seq,bahdanau2014nmt}, which generates promising performance for abstractive summarization~\cite{rush2015abs,nallapati2016seq2seq_trick}. They learn an internal language representation from a large number of examples to generate summaries similar to the ones made by humans. 

Since the used language is a critical factor of text summarization, many studies have been done for Chinese due to its large number of users, continuous use in the long history, and widespread influence in East Asia. Chinese is significantly different from any European languages especially in its character representation and word segmentation. As the old Chinese characters were developed thousands of years ago with a monosyllabic structure and used as words, the derived modern Chinese inherits a bank of characters numbering tens of thousands and composing words from either single or multiple characters. Although punctuations are widely used in modern Chinese to separate sentences, there is still no delimiter within a sentence to isolate words. Word segmentation is an error-prone process, and it largely affects the result of automatic summarization~\cite{ayana2016mrt,ma2018wean}. Since Chinese characters own semantic meanings although polysemous, many existing studies use character-based representation to simplify the effort and prevent the uncertainty of segmentation~\cite{hu2015lcsts,chen2016distraction,li2017dgrd,ayana2016mrt,ma2018wean,li2018acabs}. 

Although a few existing methods test word-based representation~\cite{hu2015lcsts,gu2016copynet}, their performances are only slightly improved or even worse. They use word-based representation on both source and target text, but it is doubtful since the approach relies on a prerequisite of sufficient training samples in terms of occurred words. However, the lengths of source and target text of a text summarization sample are definitely asymmetric. Summaries as the targets are surely shorter than their source articles and thus it is questionable whether the used dataset is large enough to provide a satisfactory size of target text. Another issue is the memory limitation. Since a Chinese word is composed of either a single or multiple Chinese characters, representing a given Chinese text dataset using words instead of characters means significantly increasing the vocabulary size. As a text summarization algorithm implemented in an encoder-decoder framework and running on a GPU for fast execution, the size of its vocabulary will be restricted by a GPU's memory capacity. To the best of our knowledge, multi-GPU platforms to expand overall memory capacity for training text summarization models are still being developed. Existing methods which use word-based representation for the decoder have to use a selected subset of the complete vocabulary bank extracted from the target text, usually the high-frequency words. However, once the size of the vocabulary subset is not large enough, many low-frequency words in the text will be replaced by the unknown token and lose their messages, which results in incomplete summaries and low ROUGE scores.

To address the problem, we propose a hybrid word-character (HWC) approach which uses hybrid embedding units for an encoder and a decoder to preserve the advantages of both word-based and character-based representations. Since an encoder does not contain a softmax layer, its computational load and memory requirement are far less than a decoder~\cite{JeanCMB14}. Thus it is feasible to apply a word-based representation on the encoder and use a large vocabulary bank. Experimental results show this approach works well on two encoder-decoder summarization methods and generates state-of-the-art performance on a widely used Chinese text summarization dataset.

\section{Related Work}
{\bf Extractive and abstractive summarization.} Numerous automatic summarization methods have been proposed in the literature, and the formats of the generated summaries categorize existing methods into two classes: extractive and abstractive. While extractive summarization selects keywords or sentences from the original text and arranges them to form a summary \cite{luhn1958sum,erkan2004lexrank,mihalcea2004textrank,cheng2016nn_extract}, abstractive summarization generates a brief version of the original text to preserve its information content ansd overall meaning. Extractive summarization is developed earlier since it highly simplifies text summarization into text partition and selection, and abstractive summarization is heavily studied recently due to its challenge and practicability. Early studies on abstractive approach include statistical machine translation techniques~\cite{banko2000statistics,knight2000statistics} and deletion-and-compression methods~\cite{cohn2008delection,filippova2015delection}. With the rapid spread of neural networks, many recent studies build their models in an encoder-decoder framework, especially the attentional Seq2Seq model~\cite{rush2015abs,nallapati2016seq2seq_trick} for its promising performance.

{\noindent \bf Chinese text summarization datasaet.} In addition to English which is the first language studied for text summarization, the large number of Chinese users motivates the studies to explore its own language features. The first compiled Chinese text dataset available for text summarization is Chinese Gigaword~\cite{ChineseGigaword}, which contains a comprehensive archive of newswire text data acquired from Chinese news sources such as Central News Agency of Taiwan and Xinhua News Agency of Beijing over several years. Although the corpus is impressive for its richness, it is neither free of charge and nor thoroughly categorized. There is a lack of human evaluation on the quality of the summaries (titles of news reports). On the contrary, the LCSTS (Large-scale Chinese Short Text Summarization) dataset~\cite{hu2015lcsts} is created for academic research. Its text sources are still news reports and titles, collected from a Chinese microblogging website. Since they are microblogging articles, their length is restricted under a short text's limit, and make the collected articles consistent. The dataset provides predefined training and test subsets, manually labeled quality indexes on its test summaries, free assess and long-term maintenance.

{\noindent \bf Open-source implementation.}
The rapid development and significant advances of many computer science fields have motivated the trend of publicly available algorithm libraries such as OpenCV~\cite{opencv_library} for computer vision, and OpenMNT~\cite{guillaume2017opennmt} for neural machine translation. Such platforms provide great convenience to test new ideas, reproduce results, and optimize performance. Since those libraries are well maintained, new methods are soon available and easily called through a unified interface. Researchers benefit from them by reducing the effort of re-implementation, preventing the errors of misunderstanding, and saving the time of conducting experiments. Due to the considerable merits, we validate the proposed HWC approach using the OpenMNT library. 

{\noindent \bf Language translation.}
Text summarization and language translation are two distinct problems in the family of natural language processing. However, they share certain similar properties such as a representation of sequential data and a conversion from text to tokens. Since the two problems are highly close, it is possible to apply models developed for one problem to the other, i.e. replacing
the source and target languages in a translation problem by the source and target text in a summarization problem. Inspired by the significant breakthrough in the translation problem~\cite{wu2016gnmt,gehring2016fairseqnmt,gehring2017fairseq,vaswani2017trans}, we apply the proposed HWC approach to a state-of-the-art translation model and find that the integrated method generates leading-edge performance.

\section{Proposed Method}
\begin{figure}[t]
  \centering
  \includegraphics[width=0.48\textwidth]{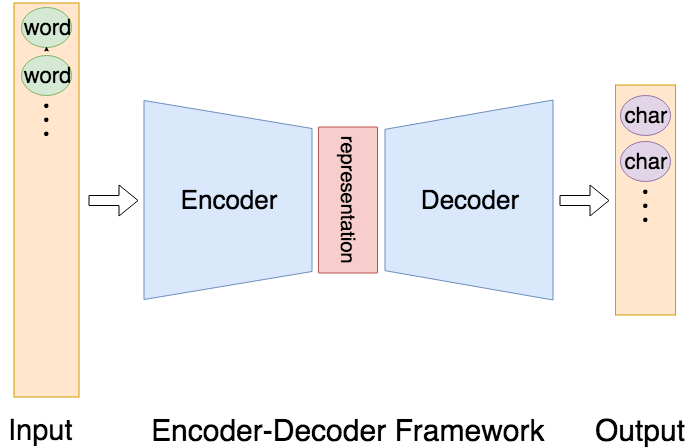}
  \caption{The input and output formats of the proposed HWC approach on an encoder-decoder summarization method.}
  \label{fig:hwc_arch}
\end{figure}

%\subsection{Model Overview}
As illustrated in Figure~\ref{fig:hwc_arch}, the proposed HWC approach is used in an encoder-decoder framework that the input articles are represented by words and the output summaries by characters. In this paper, we verify the effectiveness of the proposed HWC approach using a baseline method attentional Seq2Seq and a state-of-the-art one Transformer~\cite{vaswani2017trans}.

%\subsection{Hybrid Word-Character Model}
Different from existing methods which use a fixed type of the embedding units for both encoder and decoder as shown in Table~\ref{tab:vocab_size}, the proposed method uses word embedding to represent input articles and character embedding for output summaries. Such a design is motivated by the observation: In the Chinese language, words are more precise to provide information than characters because Chinese characters are highly polysemous, but characters are shorter and more flexible than words. On the one hand for input data, it is less ambiguous to represent articles using words than characters. In the LCSTS 1.0 dataset we use for experimental validation, there are $960k$ words made up by mere $10k$ characters. Using word embedding units expands the limit of a vocabulary bank in terms of size to train an effective encoder to capture the meaning contained in the word relationship. On the other hand for output targets, since summarized sentences are highly concentrated and it is common in Chinese to shorten long phrases for convenience, e.g. 奥运委员会~(the International Olympic Committee) is abbreviated by~奥委会~where the term~奥运~(Olympic Game) is reduced to~奥~and~委员会~(Committee) to~委会, characters are more flexible than words to represent the output sentences because of their definitely shorter length. 
\begin{table}[t]
\centering
  \resizebox{0.48\textwidth}{!}{
      \begin{tabular}{ccccc}
      \hline
      Vocabulary size & Encoder & Usage & Decoder & Usage \\ \hline
      Word-based &  961,468 & & 428,051 & \\
      \hline
      RNN+context & 10,000 & 1\% & 10,000 & 2.3\% \\
      CopyNet & 50,000 & 5\% & 50,000 & 11.6\% \\
      \hline
      \hline
      %Vocab size & Encoder & \#Usage & Decoder & \#Usage \\ \hline
      Character-based &  10,608 &  & 8,255 &  \\
      \hline
      CopyNet, DGRD, AC-ABS & 3,000 & 28.3\% & 3,000 & 36.3\% \\
      Distraction, WEAN & 4,000 & 37.7\% & 4,000 & 48.5\% \\
      \hline
      \end{tabular}
  }
  \caption{Vocabulary sizes available from the LCSTS 1.0 dataset in different representation units and the used by existing methods. The low portion of the used vocabulary in a word-based presentation tested by existing methods is caused by the limitation of GPU memory. The words are segmented from documents using the jeiba utilities. }
  \label{tab:vocab_size}
\end{table}

\section{Experimental Setup}
All of our experiments are conducted on a machine equipped with a 3.7GHz 8-core CPU, 64G memory, and a high-performance GPU (NVIDIA 1080 Ti). We use an open source implementation and the code released by the original authors to produce results of the Seq2Sqe+attn and Transformer methods respectively. We get the ROUGE scores of RNN and RNN-context methods on the LCSTS 2.0 dataset from the original authors. To train the Seq2Seq+attn and Transformer methods, we follow the existing methods RNN and CopyNet to split the Part I set into two distinct training and validation sets. Since their original authors only report random splitting rather than an explicit splitting mechanism, we use a long-standing random number generator MT19937 with five seeds (0 to 4) to select 1000 articles as the validation sets and use the remaining as training sets, and report the mean ROUGE scores.

\subsection{Dataset}
\label{Section:Dataset}
%5/21/2018 Chih-Yuan: I need to update this subsection because the abbreviation LCSTS is not first used here.
We conduct experiments on the LCSTS dataset~\cite{hu2015lcsts} to evaluate the proposed method. This dataset contains a large number of short Chinese news articles with man-made headlines as the short summaries collected from Sina Weibo\footnote{https://www.weibo.com/}, a Chinese microblogging website. This dataset is composed of three parts, as shown in Table~\ref{tab:LCSTS_data_statstic}. Part I contains a large number of 2,400,591 pairs of articles and headlines but no annotation. Parts II and III contain not only text data but also human-labeled scores, measuring the quality of summaries in terms of their relevance to the source articles. The difference between Parts II and III are the numbers of annotators to create the scores, which is 1 for Part II but 5 for Part III. The relevance scores range from 1 to 5, the larger the more relevant. 

\begin{table}[t]
\small
  \begin{center}
      \begin{tabular}{cccc}
      \hline
      Part & Version & \#Articles & \#Scores $\geq$3 \\ \hline
      \multirow{2}{*}{I} & 1.0 and 2.0 & 2,400,591 & \multirow{2}{*}{N/A} \\
      & 2.0-clean & 2,400,275 & \\
      \hline
      II & \multirow{2}{*}{1.0, 2.0, and 2.0-clean} & 10,666 & 8,685 \\
      III &  & 1,106 & 725 \\
      \hline
      \end{tabular}
  \end{center}
  \caption{The statistics of the LCSTS datasets in three different versions. The version 2.0 is updated from the version 1.0 by replacing articles in Part I which re-appear in Part III with newly collected articles. Thus the number of Part I articles does not change. The version 2.0-clean is further refined from 2.0 using a strict criterion to remove articles in Part I which are highly similar to any articles in Part III and sharing the same summaries.}
  \label{tab:LCSTS_data_statstic}
\end{table}
For a fair comparison, we follow the same split setting of existing methods \cite{hu2015lcsts,gu2016copynet,chen2016distraction,li2017dgrd,ayana2016mrt,ma2018wean,li2018acabs} to use Part I as the training set and Part III's high-relevance subset (scores equivalent to or greater than 3) as the testing set.

%\subsection*{LCSTS 2.0}
{\noindent \bf LCSTS 2.0.}
By examining the first released version (1.0) of the LCSTS dataset, we found that its Part III contains a high ratio of articles repeated in its Part I. We reported this problem to the authors, and received the response that they released an incorrect dataset which failed to filter out common articles in Parts I and III. To deal with the problem, they re-released the dataset they actually used for their experiments which replaced 329 overlapping articles in Part I with newly collected ones, and assigned it a new version 2.0.

%\subsection*{LCSTS 2.0-clean}
{\noindent \bf LCSTS 2.0-clean.}
After scrutinizing LCSTS 2.0, we find the cleanup is not complete. Many items in LCSTS 2.0's Part I are almost the same to items in Part III in terms of exactly the same summaries and highly similar articles, only differing on a few characters at the end of the articles to show the name of source newspaper. As an example shown in Table~\ref{tab:LCSTS_overlaping_example}, the name of source newspaper~新闻晨报~(Shanghai Morning Post) does not contribute to the message carried in the article. Since the issue is likely to weaken the dataset, we remove the highly repeated items from Part I and name the amended dataset LCSTS 2.0-clean.

\begin{table}[t]
\small
{\renewcommand\arraystretch{1.25}
\begin{tabular}{ll|l} 
\hline
Summary& \multicolumn{2}{l}{长时间用电子产品可能会诱发癫痫病} \\ 
\hline
\multirow{6}{*}{Article}& \multicolumn{2}{p{5.5cm}}{\raggedright 明天是第8个“国际癫痫关爱日”。最新数据显示，上海癫痫疾病发病率已达千分之八，约有10万患者。长时间使用电脑、观看电视、打游戏机等都是诱发癫痫疾病的主因。神经外科专家称，这是因为电子产品的声光刺激会致神经细胞异常兴奋。新闻晨报 } \\ 
\hline

\end{tabular}}
\caption{An example of highly overlapping items in the LCSTS 2.0 dataset. An item in the Part I split differs from another item in the Part III split only on the presence of the term~新闻晨报~(Shanghai Morning Post) at the end of the article.}
\label{tab:LCSTS_overlaping_example}
\end{table}

In order to evaluate the proposed method on a well-made dataset, we remove all highly overlapping items from the Part I split of LCSTS 2.0 and name it LCSTS 2.0-clean. The script to generate LCSTS 2.0-clean is in \href{https://github.com/playma/LCSTS2.0-clean}{Github repository} for reproducing our experimental results by other researchers.

\subsection{Evaluation Metrics}

We adopt ROUGE~\cite{lin2004rouge} metrics for evaluation, which has been widely used for abstractive summarization. They measure the quality of summaries by computing the overlap of system-generated and reference ones. For a fair comparison, we report ROUGE-1 (1-gram), ROUGE-2 (bigrams) and ROUGE-L (longest common subsequence) F1 scores for all compared methods.

\subsection{Compared Methods}

{\bf RNN and RNN-context}~\cite{hu2015lcsts} are two similar RNN-based methods except for a context generator included. In the simpler architecture without a context generator, its decoder uses the RNN encoder's last state as the input data; in the complexer architecture, a context generator is connected with all gated recurrent units' hidden states, and uses generated context to generate summaries.\\
{\bf CopyNet}~\cite{gu2016copynet} integrates the copying mechanism into the attentional Seq2Seq model in order to combine selected subsequences from the input sequence to generate an output sequence.\\
{\bf Distraction}~\cite{chen2016distraction} is a Seq2Seq framework which distracts a document into different regions by their content in order to better grasp the overall meaning of the input document.\\
{\bf DRGN}~\cite{li2017dgrd} is an attentional Seq2Seq model equipped with a latent structure modeling component.\\
{\bf MRT}~\cite{ayana2016mrt} employs the minimum risk training strategy on an attentional Seq2Seq model.\\
{\bf WEAN}~\cite{ma2018wean} is based on an attentional Seq2Seq model which generates summaries by querying distributed word representations with an attention mechanism in the decoder.\\
{\bf AC-ABS}~\cite{li2018acabs} employs an actor-critic approach originally developed for reinforcement learning on an attentional Seq2Seq model.\\
{\bf Seq2Seq+attn} is the method using the simplest attentional Seq2Seq model without any additional component. We use an implementation available from the OpenMNT system~\cite{guillaume2017opennmt} and evaluate it as a baseline.\\
{\bf Transformer}~\cite{vaswani2017trans} is a newly developed encoder-decoder method which uses attention mechanisms rather than complex recurrent or convolutional neural networks. We adopt its model for Chinese abstractive summarization and first report its performance.

\subsection{Preprocessing and Hyperparameters}
We adopt the same approach as RNN-context and CopyNet to segment input articles into words using the jieba segmentation utilities\footnote{https://pypi.python.org/pypi/jieba/}. About the hyperparameter of vocabulary size, we do experiments on several numbers as shown in Table~\ref{original_benchmark} in the same ranges used by RNN-context and CopyNet and some large numbers to assess the extent. The numbers 523506 and 523566 are the amount of words in the training sets whose occurrence frequency is greater than one. About the hyperparameters used the Seq2Seq+attn method, we empirically set the numbers of embedding dimension and hidden layers both as 500. We use Adagrad~\cite{Duchi2011Adagrad} as the optimizer, and set the initial learning rate to 0.15 and dropout rate to 0.3. We set the beam size to 5 for all decoders in our experiments. For the Transformer method, we use the default parameters of an implementation available in the OpenNMT-py repository\footnote{https://ppt.cc/ffue9x}. 

% \begin{table}[t]
% \begin{center}
%     \begin{tabular}{lrr}
%     \hline
%     vocabulary size & source & target \\ \hline
%     char-based & 10,608 & 8,255 \\ \hline
%     word-based & 961,468 & 428,051 \\
%     \hline
%     \end{tabular}
% \end{center}
% \caption{The statistics of the LCSTS 1.0 dataset's vocabulary size in different representation units.}
% \label{vocab_size}
% \end{table}

\vspace{-0.2cm}
\section{Results and Discussion}
\vspace{-0.2cm}
\begin{table*}[ht!]
  \centering
  \resizebox{1\textwidth}{!}{
  \begin{tabular}{|l|cc|cc|ccc|}
      \hline
       \multirow{2}{*}{Method} &
       \multicolumn{2}{|c}{Encoder} & \multicolumn{2}{|c}{Decoder} & \multicolumn{3}{|c|}{Measure} \\
      \cline{2-8}
      & base & vocab size & base & vocab size & ROUGE-1 & ROUGE-2 & ROUGE-L \\
      \hline
      RNN+context~\cite{hu2015lcsts} &
          char & 6000 & char & 6000 & 30.79 & N/A & N/A  \\
      \hline
      \multirow{2}{*}{CopyNet~\cite{gu2016copynet}} &
          char & 3000 & char & 3000 & 34.4 & 21.6	& 31.3 \\
          & word & 10000 & word & 10000 & 35 & 22.3 & 32 \\
      \hline
      Distraction~\cite{chen2016distraction} &
          char & 4000 & char & 4000 & 35.2 & 22.6 & 32.5  \\
      \hline
      DGRD~\cite{li2017dgrd}&
          char & 3000 & char & 3000 & 36.99 & 24.15 & 34.21 \\
      \hline
      MRT~\cite{ayana2016mrt} &
          char & 3500 & char & 3500 & 37.87 & 25.43 & 35.33 \\
      \hline
      WEAN~\cite{ma2018wean} &
          char & 4000 & char & 4000 & 37.8 & 25.6 & 35.2 \\
      \hline
      AC-ABS~\cite{li2018acabs} &
          char & 3000 & char & 3000 & 37.51 & 24.68 & 35.02 \\
      \hhline{|========|}
      Seq2Seq+attn & char & 3500 & char & 3500 &  37.25 & 24.54 & 34.32 \\
      Seq2Seq+attn (max vocab size)
      & char & 10598 & char & 8248 &  38.81 & 26.01 & 35.95 \\
      \cline{2-8} 
      HWC+Seq2Seq+attn
      & word & 50000 & char & 8248 & 40.95 & 28.58 & 38.34 \\
      HWC+Seq2Seq+attn & word & 523506 & char & 8248 & 46.17 & 33.79 & 43.62 \\
      HWC+Seq2Seq+attn (max vocab size) & word & 961212 & char & 8248 & 44.79 & 32.49 & 42.12 \\
      \hline
      Transformer& char & 3500 & char & 3500 & 42.35 & 29.38 & 39.23 \\
      Transformer (max vocab size)& char & 10598 & char & 8248 & 42.73 & 29.74 & 39.67 \\
      HWC+Transformer & word & 523506 & char & 8248 & \bf{62.26} & \bf{52.49} & \bf{60.00} \\
      \hline
  \end{tabular}
 }
  \caption{F1 scores of three ROUGE measures on the LCSTS 1.0 dataset. In the cases of maximal vocabulary size, we report the numbers of vocabularies available in training splits rather than the overall dataset (training plus validation) so that the numbers may be slightly smaller than the ones reported in Table~\ref{tab:vocab_size}. The ROUGE-1 score of the RNN+context method is given by the authors but the other two ROUGE scores are unavailable.}
  \label{original_benchmark}
\end{table*}

\begin{table*}[ht!]
  \centering
  \resizebox{1\textwidth}{!}{
  \begin{tabular}{|l|cc|cc|ccc|}
      \hline
       \multirow{2}{*}{Method} &
       \multicolumn{2}{|c}{Encoder} & \multicolumn{2}{|c}{Decoder} & \multicolumn{3}{|c|}{Measure} \\
      \cline{2-8}
      & base & vocab size & base & vocab size & ROUGE-1 & ROUGE-2 & ROUGE-L \\
      \hline
      \multirow{2}{*}{RNN~\cite{hu2015lcsts}} &
          char & 4000 & char & 4000 & 21.5 & 8.9 & 18.6 \\
          & word & 50000 & word & 50000 & 17.7 & 8.5 & 15.8 \\
      \cline{2-8}
      \multirow{2}{*}{RNN+context}
          & char & 4000 & char & 4000 & 29.9 & 17.4 & 27.2 \\
          & word & 50000 & word & 50000 & 26.8 & 16.1 & 24.1 \\
      \hline
      Seq2Seq+attn & char & 3500 & char & 3500 & 37.16 & 24.47 & 34.40 \\
      Seq2Seq+attn (max vocab size) & char & 10599 & char & 8250 &  37.42 & 24.90 & 34.81 \\
      \cline{2-8}
      HWC+Seq2Seq+attn & word & 961195 & char & 8250 & \bf{39.95} & \bf{27.40} & \bf{37.18} \\
      \hline
  \end{tabular}
  }
  \caption{F1 scores of three ROUGE measures on the LCSTS 2.0 dataset. We report the scores of the RNN and RNN+context methods here rather than in Table~\ref{original_benchmark} due to the authors' addendum as explained in Section~\ref{Section:Dataset}. }
  \label{2.0_benchmark}
\end{table*}

\begin{table*}[ht!]
  \centering
  \resizebox{1\textwidth}{!}{
  \begin{tabular}{|l|cc|cc|ccc|}
      \hline
       \multirow{2}{*}{Method} &
       \multicolumn{2}{|c}{Encoder} & \multicolumn{2}{|c}{Decoder} & \multicolumn{3}{|c|}{Measure} \\
      \cline{2-8}
      & base & vocab size & base & vocab size & ROUGE-1 & ROUGE-2 & ROUGE-L \\
      \hline
      Seq2Seq+attn & char & 3500 & char & 3500 &  35.69 & 23.24 & 33.11 \\
      Seq2Seq+attn (max vocab size) & char & 10599 & char & 8250 &  36.58 & 24.00 & 33.64 \\
      \cline{2-8}
      HWC+Seq2Seq+attn & word & 523566 & char & 8250 & 39.06 & 26.03 & 36.05 \\
      HWC+Seq2Seq+attn (max vocab size) & word & 961177 & char & 8250 & 39.41 & 26.70 & 36.58 \\
      \hline
      Transformer & char & 3500 & char & 3500 & 40.49 & 26.83 & 37.32 \\
      Transformer (max vocab size) & char & 10599 & char & 8250 & 40.72 & 27.41 & 37.43 \\
      \cline{2-8}
      HWC+Transformer & word & 523566 & char & 8250 & \bf{44.38} & \bf{32.26} & \bf{41.35} \\
      \hline
  \end{tabular}
  }
  \caption{F1 scores of three ROUGE measures on the LCSTS 2.0-clean dataset.}
  \label{2.1_benchmark}
\end{table*}
%5/21/2018 Chih-Yuan: Why is there no experiments using word representation for both encoder and decoder, we need to explain it.

%\subection{ROUGE Evaluation}
By evaluating the proposed model on the LCSTS dataset, we find that the dataset's original version (1.0) contains overlapping items and confirm the issue by its authors. Therefore, we do experiments for fair comparisons with existing methods not only on its original version, but also on two new versions (2.0 and 2.0-clean) to fairly evaluate the performance. The ROUGE scores on the three datasets are shown in Tables~\ref{original_benchmark}, \ref{2.0_benchmark} and \ref{2.1_benchmark} respectively. On the original version, the HWC+Transformer method generates the best score with a significant margin over existing methods. Using the same method on the LCSTS 2.0-clean dataset, the ROUGE scores decline reasonably and it shows the importance to evaluate methods on a well-made dataset.

%\subsection*{Vocabulary Size}
%5/18/2018 Chih-Yuan: There is a finding, but hard to reach a conclusion. Experimental results only show the case that Seq2Seq+attn (char-based) generates higher scores while using the maximal vocabulary bank. To make a conclusion, Scott need to test a series of vocabulary sizes

%%%% 5/20/2018 Scott: Transformer 的方法在 large vocabulary 上並不夠突出
{\noindent \bf Vocabulary size.}
As shown in Tables~\ref{original_benchmark}, \ref{2.0_benchmark} and \ref{2.1_benchmark}, larger vocabulary banks lead to higher ROUGE scores. On the LCTST 1.0 dataset, the char-based Seq2Seq+attn method generates higher ROUGE scores over existing methods merely by enlarging its vocabulary bank. On the LCTST 2.0 and 2.0-clean datasets, the char-based Seq2Seq+attn method also benefits from large vocabulary banks. Adopting the HWC approach, the Seq2Seq+attn method with a word-based encoder uses larger vocabulary banks and generates better ROUGE scores. On both LCSTS 1.0 and 2.0 datasets, the Transformer method generates the top performance by using the HWC approach with a large vocabulary bank.

%\subsection*{HWC}
{\noindent \bf The effects of the HWC approach.}
On all of the three datasets LCSTS 1.0, 2.0, and 2.0-clean, the HWC+Seq2Seq+attn method generates higher ROUGE scores over the Seq2Seq+attn method. Such improvements are also observed about the Transformer method on the LCSTS 1.0 and 2.0-clean dataset. Table~\ref{effect_hwc} outlines the enhancement and shows that overlapping data result in larger increases of the ROUGE scores. It indicates that the HWC+Transformer model is an aggressive learner which effectively adapts its hidden states to its training data. Thus, it generates very high ROUGE scores as the test data overlap the training ones.

\begin{table}[h]
\begin{center}
	\resizebox{0.48\textwidth}{!}{
    \begin{tabular}{lccc}
    \hline
     & ROUGE-1 & ROUGE-2 & ROUGE-L \\
    \hline
    {\bf LCSTS 1.0} & & & \\
    Seq2Seq+attn & +8.92 & +9.25 & +9.32 \\
    Transformer & +19.91 & +23.11 & +20.77 \\
    \hline
    {\bf LCSTS 2.0-clean} & & & \\
    Seq2Seq+attn & +3.72 & +3.46 & +3.47 \\
    Transformer & +3.89 & +5.43 & +4.03 \\
    \hline
    \end{tabular}
    }
\end{center}
\caption{The ROUGE scores improved by applying the HWC approach to two encoder-decoder methods.}
\label{effect_hwc}
\vspace{-0.1cm}
\end{table}

%\subsection{Analysis of Generated Summaries}
{\noindent \bf Analysis of generated summaries.}
In order to show the reason of the success caused by the HWC approach, we show four example summaries generated by the top two methods Transformer and HWC+Transformer in Table~\ref{tab:generated_example} including their source articles and man-made references. The four examples show that the HWC+Transformer method better catches the major messages contained in the source articles, which are "10 dollars" (10块钱) in Article 1, "private trusts" (私人信托) in Article 2, "23 executive meetings" (23次常务会) and "Li Keqiang" (李克强) in Article 3, and "Xian Ge Qing" (湘鄂情), "Xiaomi" (小米), and "can not compare with us" (和我们没法比) in Article 4. In contrast, the character-based Transformer method tends to copy a full sentence from its source article but miss the point. For example, it copies "I love football, but my family financial status is supported by me." (我热爱足球，但家里真的离不开我。) in Article 1 but misses the key term "10 dollars" (10块钱), "What is the highest level of the rich?" (富人的最高境界是什么?) in Article 2 but misses the key term "private trust" (私人信托), and "Combining Internet boxes such as Xiaomi and routers can not compare with us." (小米等互联网企业那些盒子和路由加一起，都和我们没法比。) but misses the key term "Xian Ge Qing" (湘鄂情). Since the proposed HWC approach helps detect key terms, it leads to higher ROUGE scores.

\begin{figure}[h]
  \centering
  \includegraphics[width=0.48\textwidth]{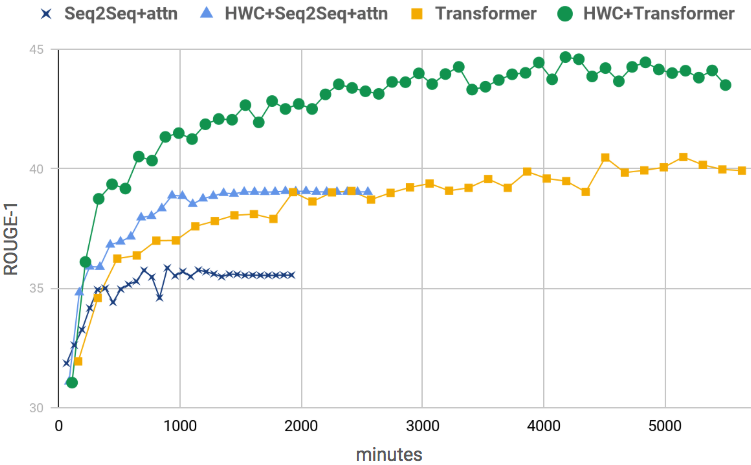} 
  \caption{Training time and performance comparison of two evaluated methods with and without the proposed HWC approach.}
  \label{fig:training_time}
	\vspace{-0.2cm}
\end{figure}

%\subsection{Training Time}
%5/21/2018 Chih-Yuan: Numbers of minutes required to train an epoch: Seq2Seq+attn: 64, HWC+Seq2Seq+attn: 85, Transformer: 161, HWC+Transformer: 110 \\ 
{\noindent \bf Training time.}
The proposed HWC approach has the advantages of not only improving performance but also increasing the training efficiency. We show the speedup as a chart in Figure~\ref{fig:training_time} from the experiments shown in Table~\ref{2.1_benchmark} conducted on the LCTST 2.0-clean dataset. Their vocabulary sizes are all of the maxima except for the HWC+Transform method due to a memory limitation. Both of the Seq2Seq+attn and Transformer methods use fewer epoches to generate high ROUGE-1 scores by applying the HWC approach. Please note the unit of the horizontal axis is a minute rather than an epoch, which means the actual execution time has been taken into consideration. For the Seq2Seq+attn method, the HWC approach increases the mean training time per epoch from 64 to 85 minutes, but for the Transformer method, it reduces the time from 161 to 110 minutes. Such result is caused by the tension between two factors that a word-based representation shortens the length of input sequences, and thus reduces the level of training difficulty, but it also increases the load of updating vocabulary embedding in the encoder due to a larger vocabulary size.

%HWC 對時間的影響可以歸納出兩個因素，其一是 HWC 縮短了 input sequence，使得訓練的難度下降。尤其在 RNN based model 之中，短的 sequence 可以較快收斂。其二是 HWC 用了較大的 vocabulary size，較大的 vocabulary size 意味著需要更多時間 update embedding。當第一個因素的影響大於第二個因素時，總體的 training time 就會下降。

\begin{table*}[!h]
\small
{\renewcommand\arraystretch{1.25}
\begin{tabular}{ll} \hline
\multicolumn{2}{p{15cm}}{\raggedright {\bf Article(1): } 睁着眼睛踢球，在亚洲杯上小组出线就能让球迷欣喜若狂；蒙起眼睛，中国曾踢到世界第二，仅次于巴西，处境却大相径庭。盲足老将林金标这次没有去参加集训：“训练虽然包吃住，但补贴每天只有10块钱。我热爱足球，但家里真的离不开我。”} \\

\multicolumn{2}{p{15cm}}{\raggedright Kick the ball when opening the eyes, fans will be ecstatic about standing out in the Asian Cup; When blindfolded the eyes, China has won the 2nd place in the world after Brazil, but the situation is very different. The blind-footed veteran Lin Jinbiao did not go to the training camp this time: “Although it covers the accommodation during the training time, the salary is only 10 dollars per day. I love football, but my family financial status is supported by me.”} \\

\\

\multicolumn{2}{p{15cm}}{\raggedright {\bf Reference: } 中国盲足曾获世界第2球员每天补贴10元钱} \\

\multicolumn{2}{p{15cm}}{\raggedright
China blind-footed player who has won the 2nd place in the world, the salary is only 10 dollars per day.}
\\
\multicolumn{2}{p{15cm}}{\raggedright {\bf Transformer: } 林金标：我热爱足球但家里离不开我} \\

\multicolumn{2}{p{15cm}}{\raggedright
Lin Jinbiao: I love football but my family financial status is supported by me.} \\

\multicolumn{2}{p{15cm}}{\raggedright {\bf HWC+Transformer: } 国足老将不参加集训只有10块钱} \\

\multicolumn{2}{p{15cm}}{\raggedright
The football veteran in China will not go to the training camp, because the salary is only 10 dollars.} \\

\hline

\multicolumn{2}{p{15cm}}{\raggedright {\bf Article(2): } 富人的最高境界是什么?钱不在自己的名下，但能自由支配。其实，私人信托最大的特点就是可以跨越生命的周期，帮助客户进行终身的个人财富管理，以及跨代的财富传承和分配，其独特的私密性、隐蔽性、稳定性更是国内很多富裕人士和家族企业的追求。} \\

\multicolumn{2}{p{15cm}}{\raggedright
What is the highest level of the rich? Money is not in their share, but it can be controlled by themselves. In fact, the biggest characteristic of private trusts is that they can cross life cycles, and help customers manage personal wealth for life, inherit and distribute wealth across generations. Their unique privacy, secrecy, and stability are among many wealthy individuals and family business pursuit in China. } \\

\\
\multicolumn{2}{p{15cm}}{\raggedright {\bf Reference: } 私人信托打造财富传承的新桥梁} \\

\multicolumn{2}{p{15cm}}{\raggedright
Private trusts create a new bridge for wealth inheritance. } \\

\multicolumn{2}{p{15cm}}{\raggedright {\bf Transformer: } 富人的最高境界是什么？} \\

\multicolumn{2}{p{15cm}}{\raggedright What is the highest level of the rich?}\\

\multicolumn{2}{p{15cm}}{\raggedright {\bf HWC+Transformer: } 私人信托的跨代生命周期} \\

\multicolumn{2}{p{15cm}}{\raggedright Private trusts cross life cycles. } \\

\hline

\multicolumn{2}{p{15cm}}{\raggedright {\bf Article(3): } 除了出访或是参加重要活动，李克强总理都会在周三主持召开国务院常务会议。7个月，23次常务会议（20次在星期三），如果将这些会议的主题用一条红线串起来，看到的不只是大政方针变化的轨迹，更有本届政府的执政之道、治国之策。} \\

\multicolumn{2}{p{15cm}}{\raggedright 
Besides visiting or participating important events, Premier Li Keqiang hosts the State Council executive meeting on Wednesday. There are 23 executive meetings (20 times on Wednesday) in 7 months. If the subject of these meetings are strung together with a red line, it shows not only the trajectory of changes in major policies and policies, but also the governance and policy of the current government.} \\

\\
\multicolumn{2}{p{15cm}}{\raggedright {\bf Reference: } 23次常务会议透视李克强执政之道} \\

\multicolumn{2}{p{15cm}}{\raggedright Analyze Li Keqiang's way of governance by 23 executive meetings. } \\

\multicolumn{2}{p{15cm}}{\raggedright {\bf Transformer: } 国务院常务会议的红线串起来} \\

\multicolumn{2}{p{15cm}}{\raggedright State Council executive meetings are strung together with a red line. } \\

\multicolumn{2}{p{15cm}}{\raggedright {\bf HWC+Transformer: } 解读李克强7个月23次常务会议} \\

\multicolumn{2}{p{15cm}}{\raggedright 
Analyze Li Keqiang's 23 executive meetings in 7 months. } \\

\hline

\multicolumn{2}{p{15cm}}{\raggedright {\bf Article(4): } 对即将改名“中科云网”的湘鄂情，孟凯充满期待，“外界说我们是做大数据玩概念，但我相信升级改造广电网络的工作，将会相当于给广电配上核武器。”“小米等互联网企业那些盒子和路由加一起，都和我们没法比。”} \\

\multicolumn{2}{p{15cm}}{\raggedright For Xian Ge Qing which is about to rename to "Zhongke Cloud Network", Meng Kai is filled with expectation. "Others say that we work on big data and its concept, but I believe that the upgrading of the radio and TV network will be equivalent to combining radio and TV with nuclear weapons. "Combining Internet boxes such as Xiaomi and routers can not compare with us."} \\

\\

\multicolumn{2}{p{15cm}}{\raggedright {\bf Reference: } 湘鄂情搞有线电视：小米们加一起都和我们没法比} \\
\multicolumn{2}{p{15cm}}{\raggedright Xian Ge Qing works on cable TV: Combining Xiaomi’s products can not compare with us.} \\

\multicolumn{2}{p{15cm}}{\raggedright {\bf Transformer: } 孟凯：互联网盒子和路由加一起都没法比} \\
\multicolumn{2}{p{15cm}}{\raggedright Meng Kai: Combining Internet and routers can not compare with us.} \\

\multicolumn{2}{p{15cm}}{\raggedright {\bf HWC+Transformer: } 湘鄂情董事长谈升级改造广电网络：小米和路由没法比} \\
\multicolumn{2}{p{15cm}}{\raggedright Chairman of Xian Ge Qing talks about upgrading radio and TV networks: Xiaomi and routers can not compare with us.} \\

\hline

\end{tabular}}
\caption{Example summaries generated from the LCSTS2.0-clean dataset.}
\label{tab:generated_example}
\end{table*}

%\begin{table}[h]
%\small
%\begin{center}
%    \begin{tabular}{l|c|c}
%    \hline
%     & Training set & Testing set \\ \hline
%    without HWC & 100.32 & 108.1 \\ 
%    HWC & 57.35 & 62.45 \\
%    \hline
%    \end{tabular}
%\end{center}
%\caption{Sequence length of input comparison between regular model and HWC on LCSTS2.1-clean dataset. }
%\label{tbe:seq_len}
%\end{table}

\section{Conclusion and Future Study}
In this paper, we propose a hybrid representation approach to improve the performance of text summarization methods in an encoder-decoder framework. Experimental results demonstrate that the proposed approach clearly generates state-of-the-art performance. In addition, we find a few errors in a widely used dataset, and provide a script to polish it. Beyond the improved performance, it may be a better representation by incorporating part-of-speech tagging into the proposed approach. We are also interested in the applicability of the proposed approach to other natural language processing problems such as dialogue generation and machine translation.

{
    %\small
    \medskip
	\bibliographystyle{acl}
	\bibliography{acl2014}

\begin{thebibliography}{}

\bibitem[\protect\citename{Ayana \bgroup et al.\egroup }2016]{ayana2016mrt}
Ayana, Shiqi Shen, Zhiyuan Liu, and Maosong Sun.
\newblock 2016.
\newblock Neural headline generation with minimum risk training.
\newblock {\em arXiv preprint arXiv:1604.01904}.

\bibitem[\protect\citename{Bahdanau \bgroup et al.\egroup
  }2014]{bahdanau2014nmt}
Dzmitry Bahdanau, Kyunghyun Cho, and Yoshua Bengio.
\newblock 2014.
\newblock Neural machine translation by jointly learning to align and
  translate.
\newblock In {\em ICLR}.

\bibitem[\protect\citename{Banko \bgroup et al.\egroup
  }2000]{banko2000statistics}
Michele Banko, Vibhu~O. Mittal, and Michael~J. Witbrock.
\newblock 2000.
\newblock Headline generation based on statistical translation.
\newblock In {\em ACL}, pages 318--325.

\bibitem[\protect\citename{Bradski}2000]{opencv_library}
Gary Bradski.
\newblock 2000.
\newblock {The OpenCV Library}.
\newblock {\em Dr. Dobb's Journal of Software Tools}.

\bibitem[\protect\citename{Chen \bgroup et al.\egroup
  }2016]{chen2016distraction}
Qian Chen, Xiaodan Zhu, Zhenhua Ling, Si~Wei, and Hui Jiang.
\newblock 2016.
\newblock Distraction-based neural networks for modeling documents.
\newblock In {\em IJCAI}, pages 2754--2760.

\bibitem[\protect\citename{Cheng and Lapata}2016]{cheng2016nn_extract}
Jianpeng Cheng and Mirella Lapata.
\newblock 2016.
\newblock Neural summarization by extracting sentences and words.
\newblock In {\em ACL}, pages 484--494.

\bibitem[\protect\citename{Cohn and Lapata}2008]{cohn2008delection}
Trevor Cohn and Mirella Lapata.
\newblock 2008.
\newblock Sentence compression beyond word deletion.
\newblock In {\em COLING}, pages 137--144.

\bibitem[\protect\citename{Duchi \bgroup et al.\egroup }2011]{Duchi2011Adagrad}
John Duchi, Elad Hazan, and Yoram Singer.
\newblock 2011.
\newblock Adaptive subgradient methods for online learning and stochastic
  optimization.
\newblock {\em Journal of Machine Learning Research}, 12:2121–2159.

\bibitem[\protect\citename{Erkan and Radev}2004]{erkan2004lexrank}
G{\"u}nes Erkan and Dragomir~R. Radev.
\newblock 2004.
\newblock Lex{R}ank: Graph-based lexical centrality as salience in text
  summarization.
\newblock {\em Journal of Artificial Intelligence Research}, 22:457--479.

\bibitem[\protect\citename{Filippova \bgroup et al.\egroup
  }2015]{filippova2015delection}
Katja Filippova, Enrique Alfonseca, Carlos~A Colmenares, Lukasz Kaiser, and
  Oriol Vinyals.
\newblock 2015.
\newblock Sentence compression by deletion with {LSTM}s.
\newblock In {\em EMNLP}, pages 360--368.

\bibitem[\protect\citename{Gambhir and Gupta}2017]{Gambhir2017}
Mahak Gambhir and Vishal Gupta.
\newblock 2017.
\newblock Recent automatic text summarization techniques: A survey.
\newblock {\em Artificial Intelligence Review}, 47(1):1--66.

\bibitem[\protect\citename{Gehring \bgroup et al.\egroup
  }2017a]{gehring2016fairseqnmt}
Jonas Gehring, Michael Auli, David Grangier, and Yann~N Dauphin.
\newblock 2017a.
\newblock {A Convolutional Encoder Model for Neural Machine Translation}.
\newblock {\em ACL}.

\bibitem[\protect\citename{Gehring \bgroup et al.\egroup
  }2017b]{gehring2017fairseq}
Jonas Gehring, Michael Auli, David Grangier, Denis Yarats, and Yann~N. Dauphin.
\newblock 2017b.
\newblock {Convolutional Sequence to Sequence Learning}.
\newblock {\em ICML}.

\bibitem[\protect\citename{Graff and Chen}2003]{ChineseGigaword}
David Graff and Ke~Chen.
\newblock 2003.
\newblock Chinese {G}igaword.
\newblock Linguistic Data Consortium.

\bibitem[\protect\citename{Gu \bgroup et al.\egroup }2016]{gu2016copynet}
Jiatao Gu, Zhengdong Lu, Hang Li, and Victor O.~K. Li.
\newblock 2016.
\newblock Incorporating copying mechanism in sequence-to-sequence learning.
\newblock In {\em ACL}, pages 1631--1640.

\bibitem[\protect\citename{Hu \bgroup et al.\egroup }2015]{hu2015lcsts}
Baotian Hu, Qingcai Chen, and Fangze Zhu.
\newblock 2015.
\newblock {LCSTS:} {A} large scale {Chinese} short text summarization dataset.
\newblock In {\em EMNLP}, pages 1967--1972.

\bibitem[\protect\citename{Jean \bgroup et al.\egroup }2014]{JeanCMB14}
S{\'{e}}bastien Jean, Kyunghyun Cho, Roland Memisevic, and Yoshua Bengio.
\newblock 2014.
\newblock On using very large target vocabulary for neural machine translation.
\newblock {\em arXiv preprint arXiv:1412.2007}.

\bibitem[\protect\citename{Klein \bgroup et al.\egroup
  }2017]{guillaume2017opennmt}
Guillaume Klein, Yoon Kim, Yuntian Deng, Jean Senellart, and Alexander~M. Rush.
\newblock 2017.
\newblock {OpenNMT}: Open-source toolkit for neural machine translation.
\newblock In {\em ACL}, pages 67--72.

\bibitem[\protect\citename{Knight and Marcu}2000]{knight2000statistics}
Kevin Knight and Daniel Marcu.
\newblock 2000.
\newblock Statistics-based summarization-step one: Sentence compression.
\newblock In {\em AAAI/IAAI}, pages 703--710.

\bibitem[\protect\citename{Li \bgroup et al.\egroup }2017]{li2017dgrd}
Piji Li, Wai Lam, Lidong Bing, and Zihao Wang.
\newblock 2017.
\newblock Deep recurrent generative decoder for abstractive text summarization.
\newblock In {\em EMNLP}, pages 2091--2100.

\bibitem[\protect\citename{Li \bgroup et al.\egroup }2018]{li2018acabs}
Piji Li, Lidong Bing, and Wai Lam.
\newblock 2018.
\newblock Actor-critic based training framework for abstractive summarization.
\newblock {\em arXiv preprint arXiv:1803.11070}.

\bibitem[\protect\citename{Lin}2004]{lin2004rouge}
Chin-Yew Lin.
\newblock 2004.
\newblock {ROUGE}: A package for automatic evaluation of summaries.
\newblock In {\em Proceedings of the ACL-04 Workshop}, pages 74--81.

\bibitem[\protect\citename{Luhn}1958]{luhn1958sum}
Hans~Peter Luhn.
\newblock 1958.
\newblock The automatic creation of literature abstracts.
\newblock {\em IBM Journal of research and development}, 2(2):159--165.

\bibitem[\protect\citename{Ma \bgroup et al.\egroup }2018]{ma2018wean}
Shuming Ma, Xu~Sun, Wei Li, Sujian Li, Wenjie Li, and Xuancheng Ren.
\newblock 2018.
\newblock Word embedding attention network: Generating words by querying
  distributed word representations for paraphrase generation.
\newblock In {\em NAACL HLT}.

\bibitem[\protect\citename{Mihalcea and Tarau}2004]{mihalcea2004textrank}
Rada Mihalcea and Paul Tarau.
\newblock 2004.
\newblock Text{R}ank: Bringing order into text.
\newblock In {\em EMNLP}.

\bibitem[\protect\citename{Nallapati \bgroup et al.\egroup
  }2016]{nallapati2016seq2seq_trick}
Ramesh Nallapati, Bowen Zhou, Caglar Gulcehre, Bing Xiang, et~al.
\newblock 2016.
\newblock Abstractive text summarization using sequence-to-sequence {RNNs} and
  beyond.
\newblock In {\em SIGNLL}, pages 280--290.

\bibitem[\protect\citename{Rush \bgroup et al.\egroup }2015]{rush2015abs}
Alexander~M. Rush, Sumit Chopra, and Jason Weston.
\newblock 2015.
\newblock A neural attention model for abstractive sentence summarization.
\newblock In {\em EMNLP}, pages 379--389.

\bibitem[\protect\citename{Sutskever \bgroup et al.\egroup
  }2014]{sutskever2014seq2seq}
Ilya Sutskever, Oriol Vinyals, and Quoc~V. Le.
\newblock 2014.
\newblock Sequence to sequence learning with neural networks.
\newblock In {\em NIPS}, pages 3104--3112.

\bibitem[\protect\citename{Torres-Moreno}2014]{TorresMoreno2014}
Juan-Manuel Torres-Moreno.
\newblock 2014.
\newblock {\em Automatic Text Summarization}.
\newblock John Wiley \& Sons, Inc.

\bibitem[\protect\citename{Vaswani \bgroup et al.\egroup
  }2017]{vaswani2017trans}
Ashish Vaswani, Noam Shazeer, Niki Parmar, Jakob Uszkoreit, Llion Jones,
  Aidan~N Gomez, {\L}ukasz Kaiser, and Illia Polosukhin.
\newblock 2017.
\newblock Attention is all you need.
\newblock In {\em NIPS}, pages 6000--6010.

\bibitem[\protect\citename{Wu \bgroup et al.\egroup }2016]{wu2016gnmt}
Yonghui Wu, Mike Schuster, Zhifeng Chen, Quoc~V. Le, Mohammad Norouzi, Wolfgang
  Macherey, Maxim Krikun, Yuan Cao, Qin Gao, Klaus Macherey, et~al.
\newblock 2016.
\newblock Google's neural machine translation system: Bridging the gap between
  human and machine translation.
\newblock {\em arXiv preprint arXiv:1609.08144}.

\end{thebibliography}
}

\end{CJK*}
\end{document}